%% file: main.tex
\documentclass[10pt]{article} 
\usepackage{fullpage}

\input{math_commands.tex}
\newcommand{\omegav}{\boldsymbol{\omega}}
\usepackage[ruled,vlined,linesnumbered]{algorithm2e}

\usepackage[utf8]{inputenc} 
\usepackage[T1]{fontenc}    
\usepackage[colorlinks,linkcolor=black,citecolor=black,urlcolor=black,bookmarks=true]{hyperref}
\usepackage{url}            
\usepackage{framed}
\usepackage{booktabs}       
\usepackage{amsfonts}       
\usepackage{nicefrac}  
\usepackage{float}
\usepackage{microtype} 
\usepackage[table,x11names]{xcolor}
\usepackage{xcolor}         
\usepackage{multirow}
\usepackage{float}
\usepackage{graphicx}
\usepackage{cleveref}
\usepackage{lipsum}

\newcommand{\blue}[1]{\textcolor{black}{#1}}

\newcommand{\teal}[1]{\textcolor{black}{#1}}
\newcommand{\circled}[1]{\raisebox{.5pt}{\textcircled{\raisebox{-.9pt} {#1}}}}

\begin{document}

\title{Indiscriminate Data Poisoning Attacks on Pre-trained Feature Extractors
\thanks{GK and YY are listed in alphabetical order.

Published in IEEE SaTML 2024. Copyright 2024 by the authors. }}


\author{Yiwei Lu, Matthew Y.R.\ Yang, Gautam Kamath, Yaoliang Yu\\
\texttt{\{yiwei.lu, m259yang, gckamath,  yaoliang.yu\}@uwaterloo.ca} \\
University of Waterloo\\
}



%
\maketitle


\begin{abstract}

Machine learning models have achieved great success in supervised learning tasks for end-to-end training, which requires a large amount of labeled data that is not always feasible. Recently, many practitioners have shifted to self-supervised learning (SSL) methods (e.g., contrastive learning) that utilize cheap unlabeled data to learn a general feature extractor via pre-training, which can be further applied to personalized downstream tasks by simply training an additional linear layer with limited labeled data. However, such a process may also raise concerns regarding data poisoning attacks. For instance, indiscriminate data poisoning attacks, which aim to decrease model utility by injecting a small number of poisoned data into the training set, pose a security risk to machine learning models, but have only been studied for end-to-end supervised learning. In this paper, we extend the exploration of the threat of indiscriminate attacks on downstream tasks that apply pre-trained feature extractors. Specifically, we propose two types of attacks:
 (1) \emph{the input space attacks}, where we modify existing attacks (e.g., TGDA attack and GC attack) to directly craft poisoned data in the input space. However, due to the difficulty of optimization under constraints, we further propose (2) \emph{the feature targeted attacks}, where we mitigate the challenge with three stages, firstly acquiring target parameters for the linear head; secondly finding poisoned features by treating the learned feature representations as a dataset; and thirdly inverting the poisoned features back to the input space. Our experiments examine such attacks in popular downstream tasks of fine-tuning on the same dataset and transfer learning that considers domain adaptation. Empirical results reveal that  transfer learning is more vulnerable to our attacks. Additionally, input space attacks are a strong threat if no countermeasures are posed, but are otherwise weaker than feature targeted attacks. 
\end{abstract}

\input{sections/intro}

\input{sections/background}

\input{sections/input_space}

\input{sections/feature}
\input{sections/experiments}

\section{Conclusions and Limitations}

In this work, we study indiscriminate data poisoning attacks on pre-trained feature extractors with different budgets, from injecting a small amount of poisoned data to perturbing the entire training set. Specifically, we consider the model parameters acquired by contrastive learning methods and poisoning the downstream task with a trainable linear head and the fixed feature extractor.
Based on the attack scheme, we propose two types of indiscriminate attacks: (1) input space attacks, including TGDA, GC, and UE input space attacks, where the attacker perturbs the feature representations indirectly through the input space, and (2) feature-targeted attacks, including decoder inversion and feature matching attacks, where an adversary acquires poisoned features and inverts them back to the input space. 

We further examine our attacks on popular downstream tasks of fine-tuning and transfer learning. Our experimental results reveal that for injecting limited numbers of poisoned samples, input space attacks generate out-of-distribution poisoned samples without constraints and become less effective after projection. On the other hand, feature targeted attacks, especially feature matching attacks exhibit an improvement with a tunable trade-off between the legitimacy of the poisoned samples and their effectiveness. Moreover, considering a more challenging domain adaptation scenario, transfer learning is more vulnerable than fine-tuning on these attacks in general.
Additionally, empirical results on unlearnable examples demonstrate that EMN is much less effective than it performs on end-to-end training with random initialization. 

\vskip 0.2cm

\noindent \textbf{Limitations and Future directions:} In this paper, we set baselines on indiscriminate attacks on pre-trained feature extractors and we believe it opens several interesting future directions: (1) Understanding the difficulty of feature inversion: prior works on adversarial examples show that small perturbation in the input space could lead to dramatic changes in the feature space for a neural network with large Lipschitz constant. In this paper, we show that a relatively small perturbation in the feature space cannot be easily inverted to the input space with constraints. Such observations might be counterintuitive and surprising, and deserve more study; (2) Possible defenses against Unlearnable Examples: although UE, e.g., EMN poses a strong indiscriminate attack, its countermeasure has not been explored. Notably, data sanitization cannot be easily performed as the entire training distribution is altered. In this paper, we provide evidence that UE is fragile against fixed feature extractors and a well-trained initialization.  Based on this, there might exist a natural defense mechanism against UE, by carefully designing the architecture or the training schemes, e.g., adding a fixed encoder for any task of interest.

\newpage\clearpage
\section*{Acknowledgments}
We thank the reviewers and program chairs for the critical comments that have largely improved the presentation and precision of this paper.
We gratefully acknowledge funding support from NSERC and the Canada CIFAR AI Chairs program. 
Resources used in preparing this research were provided, in part, by the Province of Ontario, the Government of Canada through CIFAR, and companies sponsoring the Vector Institute.

\printbibliography[segment=0]

  \newpage
  
  \clearpage
  
  \appendix

\input{rebuttal}

\newpage
\clearpage

\end{document}

%% file: math_commands.tex
\usepackage{xspace}

\newcommand{\wrt}{{w.r.t.}\xspace}

\usepackage{amsthm}

\usepackage{amsmath,amssymb,dsfont}
\DeclareMathOperator*{\argmax}{arg\,max}

\def\1{\bm{1}}

\DeclareMathAlphabet{\mathsfit}{\encodingdefault}{\sfdefault}{m}{sl}
\SetMathAlphabet{\mathsfit}{bold}{\encodingdefault}{\sfdefault}{bx}{n}

\newcommand{\zero}{\mathbf{0}}

\DeclareMathOperator*{\argmin}{arg\,min}

\newcommand{\Dsf}{\mathsf{D}}

\newcommand{\wv}{\mathbf{w}}
\newcommand{\xv}{\mathbf{x}}

\newcommand{\zv}{\mathbf{z}}

\newcommand{\gv}{\mathbf{g}}

\usepackage[english]{babel}
\usepackage[utf8]{inputenc}
\usepackage{csquotes}
\usepackage[backend=biber,style=authoryear-comp,giveninits=true,uniquename=false,maxbibnames=9,uniquelist=false,maxcitenames=2,defernumbers=true,url=false,doi=false,isbn=false]{biblatex}

\renewbibmacro{in:}{%
	\ifentrytype{article}{}{%
		\printtext{\bibstring{in}\intitlepunct}}}
\renewbibmacro*{volume+number+eid}{%
	\printfield{volume}%
	\setunit*{\addnbspace}
	\printfield{number}%
	\setunit{\addcomma\space}%
	\printfield{eid}}
\renewbibmacro*{volume+number+eid}{%
	\setunit*{\addcomma\space}
	\printfield{volume}%
	\setunit*{\addcomma\space}
	\printfield{number}%
	\setunit{\addcomma\space}%
	\printfield{eid}
}
\DeclareFieldFormat[article]{volume}{\bibstring{volume}~#1}
\DeclareFieldFormat[article]{number}{\bibstring{number}~#1}

\newbibmacro{string+doiurlisbn}[1]{%
  \iffieldundef{doi}{%
    \iffieldundef{url}{%
      \iffieldundef{isbn}{%
        \iffieldundef{issn}{%
          #1%
        }{%
          \href{http://books.google.com/books?vid=ISSN\thefield{issn}}{#1}%
        }%
      }{%
        \href{http://books.google.com/books?vid=ISBN\thefield{isbn}}{#1}%
      }%
    }{%
      \href{\thefield{url}}{#1}%
    }%
  }{%
    \href{http://dx.doi.org/\thefield{doi}}{#1}%
  }%
}
\DeclareFieldFormat*{title}{\usebibmacro{string+doiurlisbn}{\mkbibquote{#1}}} 

%% file: sections/intro.tex
\section{Introduction}
\label{sec:intro}

Modern machine learning models, especially deep neural networks, often train on a large amount of data to achieve superb performances. To collect such large-scale datasets, practitioners usually extract the desired data by crawling on the internet (e.g., web pages using Common Crawl\footnote{\url{https://commoncrawl.org/}}). However, using outsourced data 
raises an imminent security risk \parencite{NelsonBCJRSSTX08,SzegedyZSBEGF13,KumarNLMGCSX20}, namely that by carefully crafting a small amount of ``poisoned'' data, an adversary can throttle the training and hence prediction of a machine learning pipeline maliciously ~\parencite{GaoBBGHFPHTN20,Wakefield16, ShejwalkarHKR21,LyuYY20}. More formally, such a threat is called \emph{data poisoning attacks} \parencite{BiggioNL12}.

\begin{figure}[t]
    \centering
    \includegraphics[height=8.5cm]{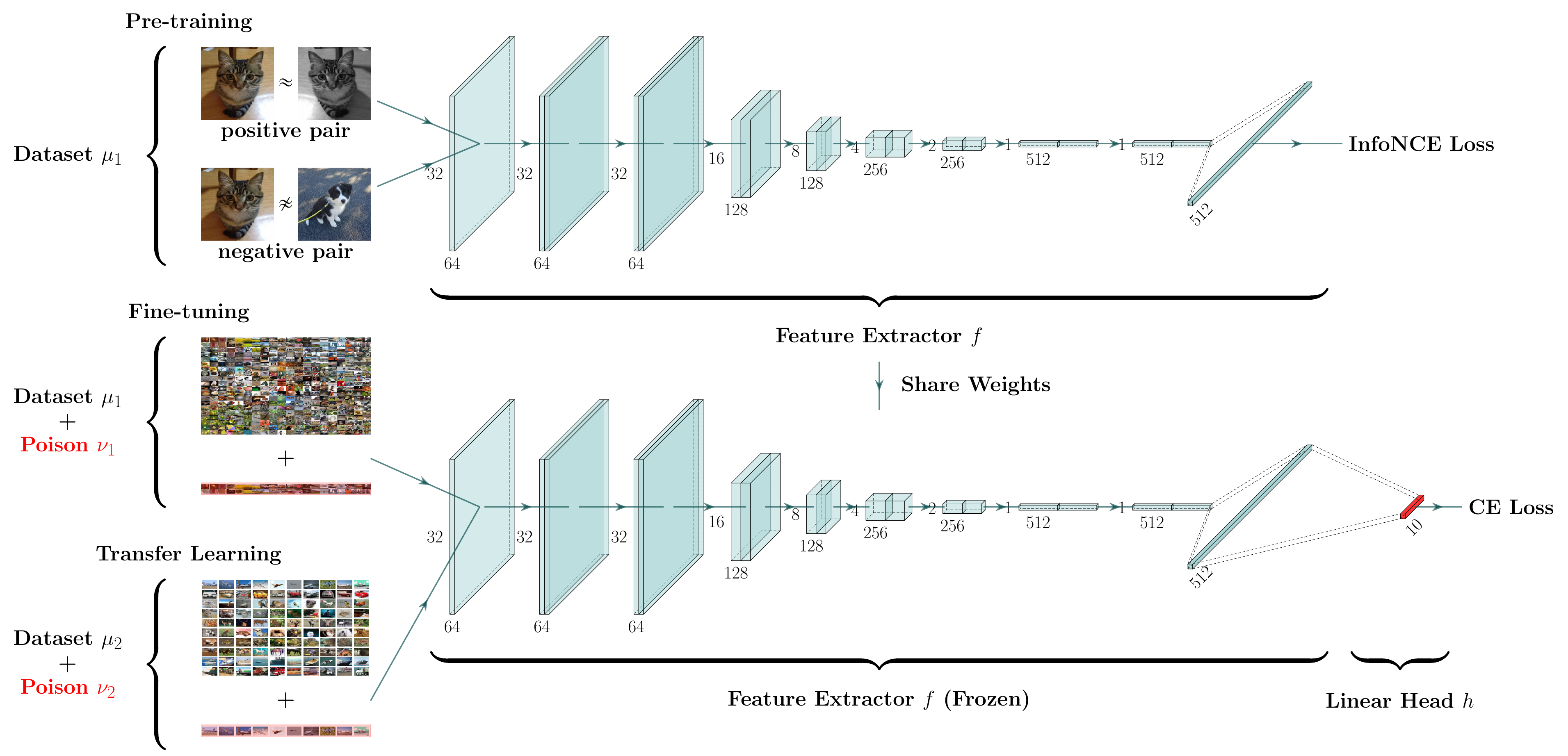}
    \caption{An illustration of our threat model:(top row) we acquire the weights of a feature extractor $f$ with contrastive learning methods and optimizing \wrt the InfoNCE loss; (bottom row) we inject poisoned samples to the training dataset on downstream applications (image classification in this example) to perturb the linear head only. We examine two scenarios in this paper: (1) fine-tuning where pre-training and downstream tasks share the same training set; and (2) transfer learning where the downstream task is performed on a different dataset. }
    \label{fig:intro}
\end{figure}

Among various data poisoning attacks (\wrt different adversary objectives), we focus on indiscriminate attacks, which aim to reduce the overall model performance during testing (e.g., accuracy for classifiers).  Indiscriminate attacks have been well-studied under end-to-end training schemes \autocite[and the references therein]{BiggioNL12,KohSL18, GonzalezBDPWLR17, SuyaMSET21, LuKY22, LuKY23}, e.g., supervised learning for image classification tasks. In particular, \textcite{LuKY23} discussed the underlying difficulties of indiscriminate attacks and proposed one of the most effective attack algorithms called Gradient Canceling (GC). 

However, as end-to-end training urges the need for large labeled datasets, practitioners with limited budgets are gradually adopting pre-trained feature extractors returned by self-supervised learning (SSL, e.g., contrastive learning methods \parencite{ChenKNH20,HeFWXG20}) for personalized downstream tasks. Specifically, SSL pre-trains a general feature extractor $f$ with potentially unlimited unlabeled data (e.g., in the wild) such that one only needs to train an additional linear head $h$ on top of $f$. While several works have studied how to poison the pre-training process on obtaining $f$ \parencite{CarliniT21,BalcanBHS22,HeZK22}, there lacks an exploration of the vulnerability of performing downstream tasks given a certified $f$ under data poisoning attacks, which could pose a serious threat to SSL beneficiaries.

Concretely, we aim to answer the following question:
\begin{quotation}
\it Can we poison downstream tasks with a fixed (and trustworthy) pre-trained feature extractor?
\end{quotation}

Specifically, we consider two popular downstream tasks in this paper: (1) use the same (but labeled) dataset as pre-training for \emph{fine-tuning}\footnote{The term ``fine-tuning'' can be used in different contexts for training with a pre-trained network as initialization. In this paper, we restrict ``fine-tuning'' to performing contrastive learning and downstream tasks on the same dataset.} the linear head $h$ (where the parameters of the feature extractor $f$ is frozen); (2) adapt to a new dataset for \emph{transfer learning}.

To address the above question, we first adopt existing data poisoning attacks in a straightforward manner (which we refer to as \emph{input space attack}) to directly craft poisoned points using the GC or TGDA attack by optimizing towards a fixed encoder. For example, considering the classification problem and the GC attack (a model-targeted approach), we perform parameter corruption only on the linear head instead of the entire model to acquire target parameters.

In practice, without imposing any constraints on the poisoned samples, we observe that GC input space attack is indeed effective (e.g., induce 29.54\% accuracy drop for fine-tuning and 39.35\% for transfer learning with only $\epsilon_d=3\%$ poisoning budget). However, such an attack can also be easily defended by data sanitization methods as the poisoned points have a much bigger magnitude and are visually malicious (see \Cref{fig:example_image}). To generate ``benign'' poisoned images, we further pose a fidelity constraint during optimization and observe the effectiveness of the attacks is largely restricted (e.g., only 2.50\% accuracy drop and 4.40\% accuracy drop under the same experimental settings).

To improve the GC input space attack, we break down the problem into three stages to mitigate the optimization difficulties: (1) acquiring target parameters of the linear head; (2) obtaining poisoned features with GC feature-space attack by treating the learned feature embeddings $f(\mu)$ as the clean dataset, and directly constructing the poisoned feature set $\zeta$, such that the linear evaluation is poor after retraining on $f(\mu)+\zeta$; (3) inverting the poisoned feature $\zeta$ back to the input space to construct $\nu$ that look visually similar to clean data, where we propose two viable approaches: decoder inversion that learns an inversion $f^{-1}$ of the feature extractor, and feature matching that optimizes $\nu$ directly.

Finally, we examine another indiscriminate data poisoning attack of a different flavor, the Unlearnable Example (UE) attack. UE operates under the strong premise that an adversary is capable of perturbing the entire training set (i.e., the poisoning fraction $\epsilon_d=\infty$), albeit  the perturbation of each point is relatively small. 

In our empirical evaluation, we observe that downstream tasks with a fixed feature extractor are surprisingly robust to the popular UE approach of \textcite{HuangMEBW21}.

In summary, we make the following contributions:
\begin{itemize}
    \item We expose the threat of indiscriminate data poisoning attacks on pre-trained feature extractors and set baselines on fine-tuning and transfer learning downstream tasks, where transfer learning is generally more vulnerable to the considered attacks;
    \item We tailor existing attacks (e.g., TGDA, Gradient Canceling attack, and Unlearnable Examples) to poisoning fixed future extractors and empirically identify the scenarios where they succeed (e.g., without constraints) and fail (with constraints);
    \item We propose new attacks called the \emph{feature targeted} (FT) attack that involves three stages to alleviate the challenge of optimizing input space attacks with constraints and observe empirical improvement.
    \item Finally, we examine unlearnable examples in the context of poisoning fixed feature extractors and identify that a popular UE attack becomes much less effective on the downstream tasks. 
\end{itemize}

%% file: sections/background.tex
\section{Background}
\label{sec:background}

In this section, we introduce the background on (1) data poisoning attacks, especially indiscriminate attacks; (2) self-supervised learning (e.g., contrastive learning) methods, and relevant notations we use throughout the paper.

\subsection{Data Poisoning attacks}

Data poisoning attack is an emerging security concern regarding machine learning models. As the primary fuel of the thriving deep learning architectures, the data collection process depends highly on uncertified online data and is exposed to adversaries, who can actively inject corrupted data into dataset aggregators (e.g., online surveys, chatbots) or simply place poisoned samples online and passively waiting for scraping \parencite{GaoBBGHFPHTN20,Wakefield16, ShejwalkarHKR21,LyuYY20}. 

Formally, data poisoning attacks refer to the threat of crafting malicious training data such that machine learning models trained on it (possibly with clean training data) would return false predictions during inference. According to different objectives, there are roughly three types of data poisoning attacks: (1) targeted attacks \parencite{ShafahiHNSSDG18,AghakhaniMWKV20,GuoL20,ZhuHLTSG19e} aim at misclassifying a specific test sample; (2) backdoor attacks \parencite{GuDG17,TranLM18,ChenLLLS17,SahaSP20} that aim to misclassifying any test sample with a specific pattern; (3) indiscriminate attacks \parencite{BiggioNL12,KL17,KohSL18,GonzalezBDPWLR17,LuKY22, LuKY23} that decrease the overall test accuracy. In this paper, we focus on indiscriminate attacks.

Specifically, we will apply a SOTA indiscriminate attack method, namely the Gradient Canceling (GC) attack \parencite{LuKY23} frequently in the later sections, and we introduce it briefly here in the traditional end-to-end training setting (along with the notations for indiscriminate attacks). Let $\ell((\xv, y), \wv)$ be our loss that measures the cost of our model parameters $\wv$ on a data sample $(\xv, y)$ for supervised learning. Here we denote the (\teal{empirical}) training distribution as $\mu$, which contains a set of training samples. Indiscriminate attacks aim at constructing a poisoned (\teal{empirical})  distribution $\nu$, where $|\nu|=\epsilon_d|\mu|$, and $\epsilon_d$ is the poisoning fraction, such that $\wv$ minimizes the loss $\ell$ by training over the mixed distribution $\chi\propto\mu+\nu$. \textcite{LuKY23} relax the optimality of the minimizer $\wv$ to having vanishing (sub)gradients over the mixed distribution:
\begin{align}
\nabla_{\wv} \ell(\chi; \wv) 
\propto \nabla_{\wv} \ell(\mu; \wv) + \nabla_{\wv} \ell(\nu; \wv) = \zero, 
\end{align}
Thus, given a good target parameter $\wv$ (usually found by parameter corruption methods \parencite{SunZRLL20}), the GC attack simply solves the following problem:
\begin{align}
\label{eq:gc}
    \argmin_{\nu \in \Gamma} ~ \tfrac12\|\nabla_{\wv} \ell(\mu; \wv) + \nabla_{\wv} \ell(\nu; \wv) \|_2^2,
\end{align}
where $\Gamma$ is the feasible set that can be specified according to the attack constraints (e.g., visual similarity to real samples to throttle defenses). 

Indiscriminate attacks usually consider a scenario where the budget is small, i.e., the poisoning fraction $\epsilon_d$ is small, such as 3\% of the training set size. Nevertheless, there exists a subclass of indiscriminate attacks called Unlearnable Examples (UE) \parencite{LiuC10,HuangMEBW21,YuZCYL21,FowlGCGCG21, FowlCGGBCG21,SadovalSGGGJ22,FuHLST21} that operate under a much stronger assumption, where $\epsilon_d = \infty$, namely that the adversary can modify the entire training set directly. In this paper, we also consider this attack as the strongest threat possible to poisoning fixed feature extractors.

\subsection{Contrastive Learning}

Recently, self-supervised learning methods, in particular, contrastive learning methods have been widely applied to learn general representations without any label information for various downstream tasks. Existing approaches aim to learn a good feature extractor $f$ (parameterized by a neural network) by minimizing the distance between representations of positive samples $f(\xv^1)$ and $f(\xv^2)$ (where $\xv^1$ and $\xv^2$ are different data augmentations of a training sample $\xv$),  while maximizing that of negative samples $f(\xv^1)$ and $f(\xv^1_n)$ (where $\xv_n$ is another sample from the training set) at the same time. Existing approaches including Contrastive Predictive Coding (CPC) \parencite{VandenoordLV18}, SimCLR \parencite{ChenKNH20, ChenKSNH20}, MoCo \parencite{HeFWXG20,HuyuhKWMK20} and $f$-MICL \parencite{LuZSGY23} apply InfoNCE-based losses to enforce the contrast between positive and negative pairs. We call this process \emph{contrastive learning pre-training}. Upon deployment, given a fixed feature extractor $f$, we consider two popular schemes of performing downstream tasks in this paper: (1) fine-tuning, where we apply the same (or slightly augmented) dataset used for pre-training and fine-tune an additional linear head on top of $f$ (also known as linear evaluation). Such a linear head can be chosen to suit specific downstream tasks, e.g., classification, object detection, and semantic segmentation. In this paper, we consider classification as an example; (2) transfer learning, where we aim at training on a personalized dataset, which is usually smaller than the dataset used in pre-training (e.g., transfer from ImageNet to CIFAR-10). Again we consider linear evaluation where a linear head is learned.

%% file: sections/input_space.tex
\section{Input Space Attacks}
\label{sec:input_space}

\begin{figure*}
    \centering
    \includegraphics[height=4.7cm]{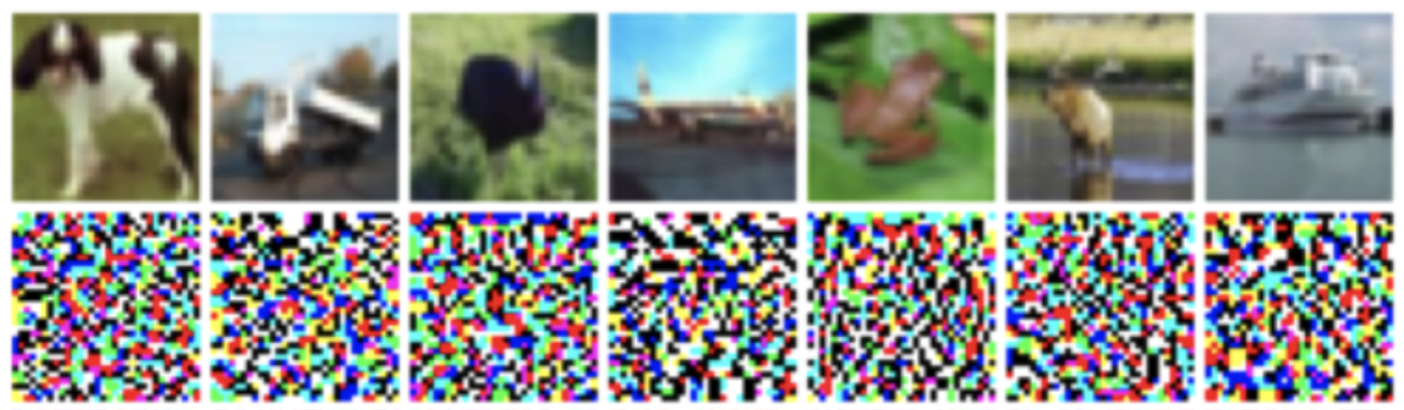}
    \caption{We visualize some clean training samples of CIFAR-10 (which serve as initialization to the attacks) in the first row, and poisoned samples generated by GC input space attacks (which induce an accuracy drop of 29.54\%) for $\epsilon_d=0.03$ in the second row. The poisoned images show that GC input space attack generates images with no semantic meaning if no explicit constraints are posed. Clean images and their corresponding poisoned ones are chosen randomly.}
    \label{fig:example_image}
\end{figure*}

Existing indiscriminate attacks on neural networks, i.e., Total Gradient Descent Ascent (TGDA) attack \parencite{LuKY22}, Gradient Canceling (GC) attack \parencite{LuKY23} and Unlearnable Examples \parencite{HuangMEBW21} generate (or optimize) poisoned points directly in the input space, which is suitable for end-to-end training. In this section, we discuss how to apply them to our problem setting, where part of the model parameters are fixed. Notably, we call this generic class \emph{input space} attacks.

\subsection{Problem Setting}

\blue{We first specify our \textbf{threat model}:
\begin{itemize}
    \item Knowledge of the attacker: the attacker has access to the training data, the target model (including its pre-trained weights), and the training procedure;
    \item Capability: the attacker can (actively or passively) inject a poisoned set to the training data; 
    \item Objective: given a pre-trained feature extractor, the objective of a downstream task and the corresponding training dataset, an adversary aims to construct a poisoned set to augment the clean training set, such that by training on the mixed dataset, the performance of the downstream task on the test set is reduced. 
\end{itemize}}

\blue{Note that in this paper we only consider indiscriminate attacks. Another strong threat is the targeted attacks, where the knowledge of the attacker is expanded to some target test data, and the objective is to alter the prediction of them. The power of such attacks on pre-trained feature extractors has been studied and we refer interested readers to \cite{ShafahiHNSSDG18}. }

We then formally introduce the problem of interest.
Given a clean training distribution $\mu$ and target model architecture that consists of a (fixed) feature extractor $f$ and a linear head $h$ with learnable parameters $\omegav$, a standard ML pipeline aims to learn a set of clean parameters $\omegav$ to minimize its cost on $\mu$ while keeping $f$ untouched:
\begin{align}
    \argmin_{\omegav} ~l(f(\mu); \omegav).
\end{align}
An adversary thus considers designing a poisoning distribution $\nu$ to maximize the cost on a clean validation set $\tilde\mu$ (which comes from the same distribution as the training set $\mu$):
\begin{align}
   \argmax_{\nu \in \Gamma} ~ l(f(\tilde\mu); \omegav).
    \label{eq:obj}
\end{align}
 As $\omegav$ is dynamically changing during optimization, solving \Cref{eq:obj} is essentially a nested optimization problem (i.e., a non-zero-sum Stackelberg game):
\begin{align}
    &\max_{\nu \in \Gamma} ~ l(f(\tilde\mu); \omegav_*), \nonumber \\
    &~\mathrm{s.t.}~ \omegav_* \in \argmin_{\omegav} ~ l(f(\mu)+f(\nu); \omegav),
    \label{eq:bilevel}
\end{align}
where the poisoning effectiveness is measured by retraining the linear head $\omegav$ on the mixed feature distribution $f(\mu)+f(\nu)$. Note that the attack budget is measured by the poisoning fraction $\epsilon_d = |\nu|/|\mu|$. Firstly, we introduce two viable approaches for solving the problem when $\epsilon_d$ is small (e.g., $<100\%$), namely the TGDA attack \parencite{LuKY22} and GC attack \parencite{LuKY23}. Next, we discuss UE \parencite{HuangMEBW21}, where $\epsilon_d = \infty$, which amounts to substituting the entire training distribution $\mu$ as $\nu$. 
Notably, we call these adaptations TGDA input space attack, GC input space attack and UE input space attack, respectively.

\subsection{TGDA input space attack}

Solving \Cref{eq:bilevel} is non-trivial, especially for the outer maximization problem as the dependence of $l(f(\tilde\mu); \omegav_*)$ on $\nu$ is indirectly through retraining the head model on the mixed distribution. As a result, one can not directly apply simple algorithms such as gradient ascent as it would lead to zero gradients. \textcite{LuKY22} propose to measure the influence of $\nu$ through the total derivative $\Dsf_{\nu} l(f(\tilde\mu); \omegav) = -\nabla_{\omegav\nu} l_2 \cdot \nabla_{\omegav\omegav}^{-1} l_2 \cdot \nabla_{\omegav} l_1$ (where we simplify $l_1 = l(f(\tilde\mu),\omegav), l_2 = l(f(\mu)+f(\nu); \omegav)$), which implicitly appraises the change of $\omegav$ with respect to $\nu$. Specifically, TGDA considers the following total gradient ascent step for the attacker and gradient descent step for the defender:
\begin{align}
    \nu &= \nu + \eta \cdot \Dsf_{\nu} l(f(\tilde\mu); \omegav), \\
    \omegav &= \omegav - \eta \cdot \nabla_{\omegav} l(f(\mu)+f(\nu); \omegav) .
\end{align}
For implementation, we still perform the attack end-to-end, namely that we directly feed the data into the model architecture, while keeping $f$ untouched (e.g., \texttt{requires\_grad=False} in PyTorch).

\begin{figure*}
    \centering
    \includegraphics[height=7.8cm]{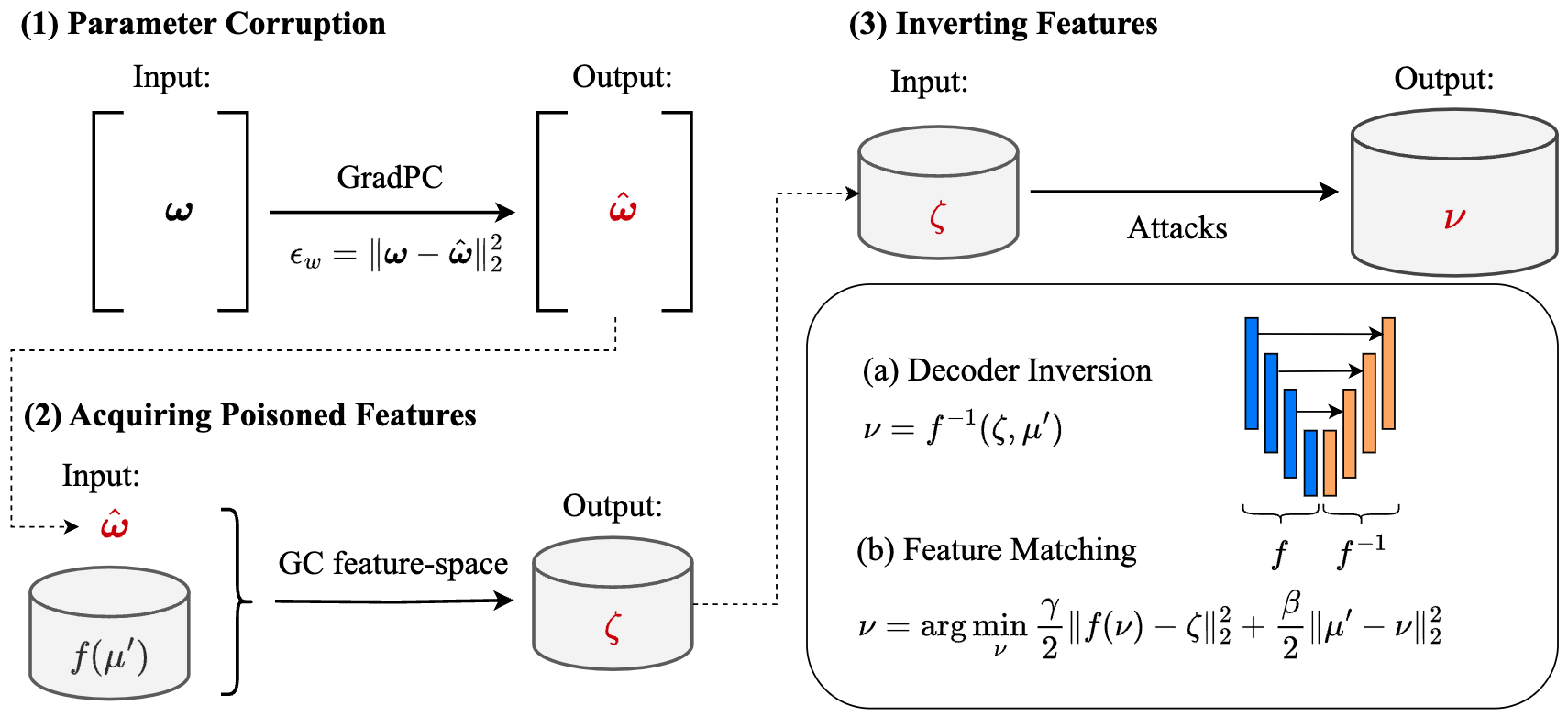}
    \caption{An illustration of the three stages of feature-targeted attacks: (1) obtaining the target linear head parameter $\omegav$ with GradPC; (2) acquiring poisoned features $\zeta$ with GC feature-space attack; (3) invert $\zeta$ back to the input space using feature matching or decoder inversion (decoder inversion requires training an autoencoder with a fixed encoder $f$).}
    \label{fig:FT_attack}
\end{figure*}

\subsection{GC input space attack}
Next, we generalize the GC attack (discussed in \Cref{sec:background}) to our problem setting. Recall that to avoid solving the difficult \Cref{eq:bilevel}, GC aims to reach a specific target parameter $\hat\omegav$ that the model will converge to by retraining on the mixed feature distribution. Here reachability is defined by having a vanishing (sub)gradient over the mixed feature distribution:
\begin{align}
\nabla_{\omegav} l(f(\chi); \hat\omegav) 
\propto \nabla_{\omegav} l(f(\mu); \hat\omegav) + \nabla_{\omegav} l(f(\nu); \hat\omegav) = \zero.
\end{align}
Thus, GC input space attack amounts to solving the following problem:
\begin{align}
\label{eq:gc_input}
    \argmin_{\nu \in \Gamma} ~ \tfrac12\|\nabla_{\omegav} l(f(\mu); \hat\omegav) + \nabla_{\omegav} l(f(\nu); \hat\omegav) \|_2^2.
\end{align}
Note that GC requires a target parameter $\hat\omegav$ that performs poorly on the test set, which  \textcite{LuKY23} acquired by another attack that perturbs the clean model parameters $\omegav$ directly, namely the Gradient Parameter Corruption (GradPC) attack of \textcite{SunZRLL20}. In a nutshell, GradPC overwrites $\omegav$ with $\hat{\omegav}$ directly in our problem setting using a gradient approach. Note that the perturbation introduced by GradPC is measured by $\epsilon_w = \|\omegav - \hat{\omegav}\|_2^2$ and can be specified as a constraint by the algorithm.

In practice, we observe that GC input space attack is very effective without considering any constraints on $\nu$, e.g., it approximately reaches the target parameter $\hat{\omegav}$ for $\epsilon_d=0.1$ in \Cref{tab:finetuning} and \Cref{tab:transfer}, \Cref{sec:exp}. However, the poisoned samples generated by GC has a significantly larger magnitude than the clean distribution $\mu$ and does not retain semantic information (see \Cref{fig:example_image} for a visualization). To further generate ``benign'' poisoned points to throttle defenses, we add a penalty term $\tfrac{1}{2} \|\mu' \!-\!\nu\|_2^2$ in \Cref{eq:gc_input} on constraining $\nu$ to be similar with $\mu'$, which is part of $\mu$ that initializes $\nu$ ($|\mu'|=|\nu|$). 

However, we find the optimization procedure to be much harder with such a constraint, and the attack effectiveness is largely restricted. For example, we observe an accuracy drop of only 2.5\% compared to 29.54\% (without constraints) in \Cref{tab:finetuning}. To mitigate the difficulty of optimizations, we further propose feature targeted attacks in \Cref{sec:feature} by breaking down the problem into several stages.

\subsection{UE (EMN) input space attack}

Before diving into \Cref{sec:feature}, we introduce the last input space attack we consider, the  Unlearnable Examples (UE), where we pose a strong assumption on $\epsilon_d=\infty$. We apply the error-minimizing noise (EMN) approach in \textcite{HuangMEBW21}.
Here the task is to perturb the entire training set with a noise factor $\delta$ such that the model learns a strong correlation between $\delta$ and the label information. Note that we consider $\delta$ to be a sample-wise noise, namely that $\mu+\delta = \{\xv_i + \delta_i\}_{i=1}^{|\mu|}$. EMN input space attack optimizes the following objective:
\begin{align}
    \min_{\omegav} ~ \min_{\delta \in \Gamma} ~ l(f(\mu+\delta);\omegav).
    \label{eq:UE}
\end{align}
Note that optimizing \Cref{eq:UE} is substantially easier than \Cref{eq:bilevel}, as it is a min-min problem and can be easily solved by (alternating projected) gradient descent algorithms.

%% file: sections/feature.tex
\section{Feature Targeted Attack}
\label{sec:feature}

Motivated by the optimization challenge of GC input space attack, we propose a staged strategy to mitigate the difficulty of the constrained problem. Assuming we \textcircled{1} acquired the target parameter $\hat{\omegav}$ with parameter corruption, let us rewrite \Cref{eq:gc_input} by introducing a feature distribution $\zeta$:
\begin{align}
\min_{\nu\in\Gamma} ~ \min_{\zeta = f(\nu)} ~ \tfrac12\| \underbrace{\nabla_{\omegav} l(f(\mu); \hat\omegav)}_{:= \gv} + \nabla_{\omegav} l(\zeta; \hat\omegav) \|_2^2.
\end{align}
Next, we relax both constraints by introducing penalty terms\footnote{Other penalty terms other than the (squared) $L_2$ norm could also be used.}: 
\begin{align}
\label{eq:gc_input-relax}
    \min_{\nu,\zeta} \underbrace{\tfrac12\|\gv \!+\! \nabla_{\omegav} l(\zeta; \hat\omegav) \|_2^2}_{\circled{2}} + \underbrace{\tfrac{\gamma}{2}\|f(\nu) \!-\! \zeta\|_2^2 + \tfrac{\beta}{2} \|\mu' \!-\!\nu\|_2^2}_{\circled{3}}.
\end{align}
The variables $\nu$ and $\zeta$ are now separated, which immediately suggests an alternating algorithm that we elaborate on below.

\begin{figure*}
    \centering
    \includegraphics[height=9cm]{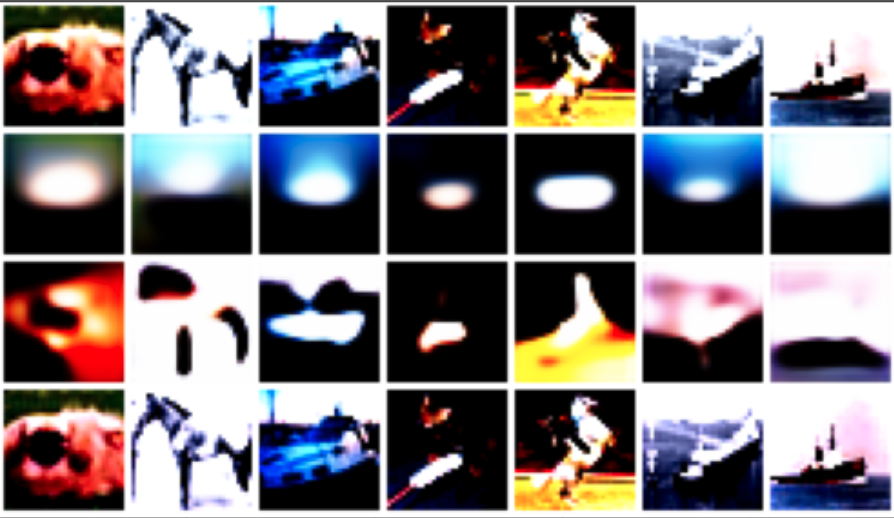}
    \caption{Here we visualize original test images (first row), and images reconstructed by an autoencoder with fixed ResNet-18 feature extractor learned by MoCo (second row), the same autoencoder trained end-to-end (third row), and the autoencoder with skip connections, i.e., a U-Net (fourth row).}
    \label{fig:u-net}
\end{figure*}

We follow the above three steps and break down the optimization into corresponding three stages: \circled{1}: acquiring the target parameter $\hat{\omegav}$ with GradPC; solving \circled{2} with GC feature space attack, where we treat the learned representation as the clean dataset and inject poisoned features directly; solving \circled{3} where we invert the poisoned features back to the input space considering constraints in the input space, where we introduce two possible approaches: decoder inversion (without optimization) and feature matching (with optimization). Overall, we call this approach \emph{Feature Targeted} (FT) attack and we visualize the attack procedures in \Cref{fig:FT_attack}.

\subsection{GC feature space attack}

Firstly, we generate poisoned features $\zeta$ using the GC attack. We chose GC as it is the best-performing algorithm in the feature space. We consider $f(\mu)$ as the clean feature training set and enable an attacker to inject poisoned features $\zeta$ directly. Following the notations in \Cref{sec:input_space}, GC considers reachability in the feature space:
\begin{align}
\nabla_{\omegav} l(f(\mu); \hat\omegav) + \nabla_{\omegav} l(\zeta; \hat\omegav) = \zero, 
\end{align}
and optimizes the following problem:
\begin{align}
\label{eq:gc_feature}
    \argmin_{\zeta} ~ \tfrac12\|\nabla_{\omegav} l(f(\mu); \hat\omegav) + \nabla_{\omegav} l(\zeta; \hat\omegav) \|_2^2,
\end{align}
which corresponds to the term $\circled{2}$ in \eqref{eq:gc_input-relax}.
We denote $\zeta$ as the output of the GC feature space attack and apply it as input to the next stage.

Here the target parameter $\hat\omegav$ is generated by \circled{1} GradPC from the feature space, where we treat $f(\mu)$ as the training set and perturb the linear head $h$. Unlike poisoning attacks, such a change does not make a significant difference for GradPC as it perturbs the linear head parameters directly.

\subsection{Inverting features}

Next, we show two viable approaches to solve \circled{3} on inverting the poisoned feature back to the input space.

\subsubsection{Decoder Inversion}

First, we show how to solve \circled{3} without optimizing $\nu$ directly by training a decoder network. Specifically,
given a pre-trained feature encoder $f$ that maps $\mu$ to $f(\mu)$, we aim to learn its inversion $f^{-1}$ such that $f^{-1}(f(\mu)) \approx \mu$. For image reconstruction and related tasks (e.g., image inpainting, image-to-image translation, semantic segmentation), it is common to utilize an autoencoder architecture, where the encoder is $f$ and the decoder is naturally $f^{-1}$. However, such an autoencoder is usually trained end-to-end, while we require training only the decoder $f^{-1}$ with a fixed encoder $f$. In \Cref{fig:u-net} (rows 1-3), we apply a ResNet-18 encoder learned by MoCo on CIFAR-10 and train the corresponding decoder on reconstructing CIFAR-10. We show such a constraint indeed brings a challenge: the autoencoder with a fixed $f$ does not learn the detailed structure of images compared to that trained end-to-end, which may imply that an encoder trained by contrastive learning is not necessarily a good encoder for image reconstruction. Additionally, we observe that a plain autoencoder does not reconstruct the images well, even when training end-to-end.

Previous works on image reconstruction tasks show that a key success of the architectural design is to add skip connections between encoders and decoders, also known as U-Net \parencite{RonnebergerFB15}. In \Cref{fig:u-net} (row 4), we show that adding skip connections leads to superb reconstruction quality, even with a fixed encoder architecture. 

Upon acquiring a good decoder $f^{-1}$, we can then invert the poisoned feature $\zeta$ back to the input space. Note that with a U-Net architecture, $f^{-1}$ does not only take $\zeta$ as input but also the intermediate features of $\mu'$ \footnote{Here $\mu'$ refers to the clean samples corresponding to $f(\mu')$, which are the initialization of the GC feature space attack.}, which we denote as $f_{1}^n(\mu')$, where $n$ is the number of skip connections. In summary, the output of the decoder inversion is $\nu = f^{-1}(\zeta, f_{1}^n(\mu'))$, where the first term in \circled{3} is minimized if the decoder is well-trained, and the second term is automatically handled by the skip connections.

\subsubsection{Feature Matching}

In decoder inversion, we train the decoder $f^{-1}$ and hoping the acquired $\nu$ is a successful inversion, i.e., $f(\nu) = \zeta$. However, in practice, we find the distance between the two is still large, e.g., $\|f(\nu)-\zeta\|_2^2\approx 150$, which indicates the gap between the decoder inversion and the implicit ground truth is still large (another indication is the performance gap between GC feature space attack and decoder inversion attack in \Cref{sec:exp}).

To further close the gap between $f(\nu)$ and $\zeta$,  we introduce an alternative method that directly optimizes the poisoned dataset $\nu$ by matching its feature $f(\nu)$ to the target poisoned feature $\zeta$ while enforcing the distance between $\mu$ and $\nu$ to be small. We call this algorithm \emph{Feature Matching}.

Our feature matching algorithm is inspired by the feature collision (FC) algorithm \parencite{ShafahiHNSSDG18} in targeted attack. In targeted attacks, FC considers a target test sample $t$ in the test set and a different base instance $b$. FC aims to construct a single poisoned data sample $\xv_p$ such that the distance between $f(t)$ and $f(\xv_p)$ in the feature space is small while $\xv_p$ is close to $b$ in the input space. 

When adapting to our setting, instead of matching the feature of a target sample, we match the target feature $\zeta$ that we already acquired. Thus, we need only optimize the term \circled{3} in \eqref{eq:gc_input-relax}:
\begin{align}
    \argmin_{\nu} ~ \tfrac{\gamma}{2}\|f(\nu) - \zeta\|_2^2 + \tfrac{\beta}{2}\|\mu' - \nu\|_2^2,
\end{align}
where recall that the second term is a relaxation of the constraint $\nu\in\Gamma$. 
We present our algorithm in \Cref{alg:FT}, where we follow \textcite{ShafahiHNSSDG18} and apply a forward gradient step to minimize the first term on feature matching and a backward proximal step to enforce the constraint on the input space. Note that $\beta$ is a hyperparameter for controlling the strength of the constraint.    

Again, $\mu'$ is part of the training distribution where $|\mu'|=\epsilon_d|\mu|=|\zeta|$. Instead of choosing $\mu'$ from the initialization of $\zeta$, as in decoder inversion, 
we follow the following better strategy. 
The key observation is that in practice, $\zeta$ is not necessarily close to the initial corresponding $\mu'=\nu_0$, thus making the initial optimization hard. In other words, we find it necessary to explore whether the choice of $\mu'$ can be optimized in order to find a better matching in the input space. A simple solution is to perform a grid search for individual poisoned features: for every $\zv \in \zeta$, we calculate the point-wise distance $\|f(\xv) - \zv\|_2^2$ for every $\xv \in \mu$. By ranking the distances, we simply choose the one with the smallest distance to form a pair $\{\xv, \zv\}$. Such a grid search provides a much better initialization and significantly reduces the running time upon convergence. Moreover,  we find that such a process reduces the difficulty of optimization and helps the algorithm converge at a lower value.

\begin{algorithm}[t]
\DontPrintSemicolon
    \KwIn{base instances as part of the training distribution $\mu' = \epsilon_d\mu$,  
    step size $\eta$, a pre-trained encoder $f$} 
    \KwOut{poisoned dataset $\nu$}
    
    $\nu \gets \mu'$ \tcp*{initialize poisoned data}

    $\zeta \gets f(\nu)$ \tcp*{initialize poisoned feature}
    \For{$k=1, ...$}{

        \For{$s=1, 2, ...$}{
        $\zeta \gets \zeta - \eta \cdot 
        \nabla_{\zeta} [\tfrac12\|\gv + \nabla_{\omegav} l(\zeta; \hat\omegav) \|_2^2]$ 
        
        $\zeta \gets [\zeta + \eta\gamma f(\nu)] / (1+\eta\gamma)$ \tcp*{optional}
        }
        
        \For{$t =1, 2, ...$}{
        $\nu \gets \nu - \eta \cdot\nabla_{\nu} [ \tfrac{\gamma}{2} \|f(\nu) - \zeta\|_2^2]$ \tcp*{forward}

        $\nu \gets (\nu+\eta\beta\mu') / (1+\eta\beta)$ \tcp*{backward}
        }
    }
\caption{Feature Matching}
\label{alg:FT}
\end{algorithm}

%% file: sections/experiments.tex
\section{Experiments}
\label{sec:exp}

\begin{table*}[ht]
    \centering
    \caption{Fine-tuning results: comparison between input space attacks (TGDA and GC input space attacks with/without constraints on poisoned samples) and feature targeted attacks (Decoder inversion and Feature Matching with different $\beta$) \wrt different $\epsilon_d$. \teal{The white background indicates methods that generate target parameters or features; blue indicates input space attacks and pink indicates feature targeted attacks.}}
    \label{tab:finetuning}  
    \setlength\tabcolsep{6pt}
   \resizebox{\textwidth}{!}{\begin{tabular}{lrrrrrr}
\toprule
\bf Attack method & Clean & $\epsilon_d=0.03$ \teal{(1500)} & $\epsilon_d=0.1$ \teal{(5000)} & 
$\epsilon_d=0.5$ \teal{(25000)} & $\epsilon_d=1$ \teal{(50000)}\\

\midrule
GradPC (input space) & 89.71\% & 53.15\% / -36.56\% & 53.15\% / -36.56\% & 53.15\% / -36.56\% & 53.15\% / -36.56\% \\

\rowcolor{LightCyan1}
TGDA-input (no constraints) & 89.71\% & 85.43\% / -\phantom{1}4.28\% & 84.26\% / -\phantom{1}5.45\% & 76.55\% / -13.16\% & 70.38\% / -19.33\% \\

\rowcolor{LightCyan1}
TGDA-input (with constraints)  & 89.71\% & 87.52\% / -\phantom{1}2.19\% & 87.45\% / -\phantom{1}2.26\% & 83.67\% / -\phantom{1}6.04\% & 78.66\% / -11.05\% \\

\rowcolor{LightCyan2}
GC-input (no constraints) & 89.71\% & 60.17\% / -29.54\% & 53.65\% / -36.06\% & 53.15\% / -36.56\%  & 53.15\% / -36.56\% \\

\rowcolor{LightCyan2}
GC-input (with constraints) &  89.71\% & 87.21\% / -\phantom{1}2.50\% & 87.05\% / -\phantom{1}2.66\% & 80.30\% / -\phantom{1}9.41\% & 75.33\% / -14.38\% \\

\midrule

GradPC (feature space) & 89.71\% & 53.00\% / -36.71\% & 53.00\% / -36.71\%  & 53.00\% / -36.71\%  & 53.00\% / -36.71\%  \\

GC-feature space & 89.71\% & 53.10\% / -36.61\% & 53.03\% / -36.68\%  & 54.09\% / -35.62\%  & 52.15\% / -37.56\%  \\

\rowcolor{LavenderBlush1}
Decoder inversion & 89.71\% & 85.49\% / -\phantom{1}4.22\%  & 84.61\% / -\phantom{1}5.10\%  & 78.65\% / -11.06\%  & 72.25\% / -17.46\% \\

\rowcolor{LavenderBlush2}
Feature matching ($\beta=0.25$)  & 89.71\% & 83.41\% / -\phantom{1}6.30\%  & 82.33\% / -\phantom{1}7.38\%  & 76.15\% / -13.56\%  & 69.17\% / -20.54\%  \\

\rowcolor{LavenderBlush2}
Feature matching  ($\beta=0.1$) & 89.71\% & 77.24\% / -12.44\% & 76.56\% / -13.15\% & 73.24\% / -15.47\% & 65.14\% / -24.57\% \\

\rowcolor{LavenderBlush2}
Feature matching  ($\beta=0.05$) & 89.71\% & 75.34\% / -14.37\% & 74.29\% / -15.42\% & 71.99\% / -17.72\% & 63.19\% / -26.52\% \\

\bottomrule
\end{tabular}}
\vspace{-10pt}
\end{table*}

In this section, we present our experimental results on the proposed indiscriminate attacks. For a fixed feature extractor, we consider two approaches to performing supervised downstream tasks: (1) fine-tuning, where the contrastive learning pre-training and fine-tuning share the same training distribution (the difference lies on without/with label information); (2) transfer learning, where pre-training is performed on a (relatively) large-scale dataset (without label information), and the linear head is trained on a smaller or customized dataset with labels. 

Our experiments will thus be structured as these two major parts, where we consider input space attacks and feature space attacks for each class. 

\subsection{Experimental setting}

\noindent \textbf{Hardware and Package:} \teal{Fine-tuning experiments are run on a cluster with NIVIDIA T4 and P100 GPUs, while transfer learning experiments are run on another machine with 2 NIVIDIA 4090 GPUs.} The platform we use is \texttt{PyTorch}. 

\noindent \textbf{Datasets:} We consider image classification as the downstream task on the CIFAR-10 dataset \parencite{Krishevsky09} (50k training and 10k testing images). For contrastive learning pre-training, we use CIFAR-10 and ImageNet 1K \parencite{DengDSLLF09} for fine-tuning and transfer learning respectively.  

\noindent \textbf{Feature Extractors:} Following the standard contrastive learning pre-training architectures, we examine the ResNet-18 \parencite{HeZRS16} model with a feature dimension of size 512 for pre-training on CIFAR-10, following the common recipe of changing the convolution kernel size from $7\times7$ to $3\times3$. The model weights are obtained by training on SimCLR \parencite{ChenKNH20} for 800 epochs with a batch size of 512. For clean linear evaluation, we acquire an accuracy of 89.71\% with 100 epochs. 
For ImageNet pre-training we apply the ResNet-50 architecture using MoCo V3 \parencite{ChenXH21} with a feature dimension of size 2048 and a batch size of 4096, trained for 1000 epochs. We directly utilize the pre-trained weights provided in the MoCo V3 Github Repo\footnote{\url{https://dl.fbaipublicfiles.com/moco-v3/r-50-1000ep/r-50-1000ep.pth.tar}}. Upon transfer learning, we follow the general recipe to resize the CIFAR-10 training sample from $32\times 32$ to $224\times224$ and acquire an accuracy of 93.03\% by training the linear head for 100 epochs. 

\noindent \textbf{Attacks implementations:} We specify the implementations of all attacks we apply in this paper:\\
(1) TGDA input space attack: we follow and modify the implementation of \textcite{LuKY22}\footnote{\url{https://github.com/watml/TGDA-Attack}} while fixing the feature extractor of the follower (defender) and train for 200 epochs for all tasks; \\
(2) GC input space/feature space attack: we follow and modify the official code of \textcite{LuKY23}\footnote{\url{https://github.com/watml/plim}} and perform 2000 epochs or early stop upon reaching stopping criteria (loss smaller than 1). Note that for acquiring target parameters, we run GradPC\footnote{\url{https://github.com/TobiasLee/ParamCorruption}} for one single attack step and set $\epsilon_w=1$; \\
(3) UE input space attacks: we follow and modify the implementation of \textcite{HuangMEBW21}\footnote{\url{https://github.com/HanxunH/Unlearnable-Examples}} for sample-wise error minimizing noise generation; \\
(4) Decoder inversion: for the architectural design on the decoder, we use the implementation of this Github repo\footnote{\url{https://github.com/mkisantal/backboned-unet}} as a key reference. For image reconstruction, we apply a combination of the multi-scale structural similarity index (MS-SSIM) loss \parencite{WangSB03} and $L_1$ loss following the analysis of  \cite{KTKO21}; \\
(5) Feature matching: finally, we follow \Cref{alg:FT} and set $k=1$, $t=2000$. We set the learning rate as 0.1 with a cosine scheduler. Unless specified, $\gamma=1, \beta=0.25$ across all tasks. 

\noindent \textbf{Label Information:} During optimizations of indiscriminate attacks, we only modify the input $\xv$ and generate clean-label poisoned samples. Specifically, we assign the label corresponding to the clean images during the initialization stage of every attack. Note that for feature matching, as we perform ranking after the GC feature space attack, we reassign the labels as that of clean base samples to every $\zeta$, thus the output $\nu$.

\noindent \textbf{Evaluation Protocol:} To evaluate the performance of each attack, we first acquire the poisoning dataset $\nu$ and retrain the model (initialized with the same random seed across all attacks) on both clean and poisoned data $\mu+\nu$ until convergence (we train for 100 epochs across all tasks). Our evaluation metric is the test accuracy drop, compared with clean accuracy (obtained by training on clean data only).

\subsection{Fine-tuning}
We report our main results for poisoning the fine-tuning downstream task in \Cref{tab:finetuning}. We expand discussions and our key observations in the following paragraphs for input space attacks and feature-targeted attacks, respectively.

\subsubsection{Input space attacks}

\begin{figure*}
    \centering
    \includegraphics[height=9cm]{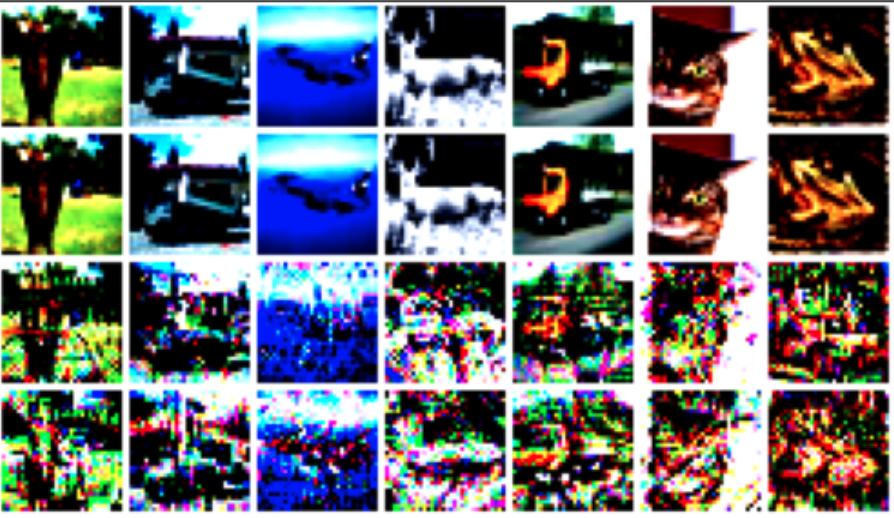}
    \vspace{-5pt}
    \caption{Here we visualize clean images (first row), and poisoned samples returned by the feature matching algorithm with $\beta=0.25, 0.1, 0.05$ respectively from the second to the fourth row.}
    \label{fig:feature_matching}
    \vspace{-10pt}
\end{figure*}

Recall in \Cref{sec:input_space}, we introduce the TGDA input space attack and the GC input space attack. We first examine the two attacks without considering any constraints, such that the poisoned points $\nu$ could be randomly large and do not need to resemble the real distribution $\mu$. Although under this setting the poisoned samples can be easily detected, it is still valuable to study as it reveals the security threats without any data sanitization process, which is not uncommon in practice. From \Cref{tab:finetuning} (upper part, row 1,2,4), we observe that compared to TGDA, GC is very effective, where it can almost reach the target parameter generated by GradPC when $\epsilon_d=0.1$. 

However, GC generates poisoned samples that have a much larger magnitude than the clean samples, especially when $\epsilon_d$ is small (e.g., $100 \times$ larger for $\epsilon_d=0.03$ and decreases to $5 \times$ larger for $\epsilon_d=1$). This also matches the observations in \textcite{LuKY23}: a higher $\epsilon_d$ is desired for an attacker as it not only increases the attack effectiveness but also brings the poisoned points closer to the clean distribution, at least magnitude-wise.

Additionally, \textcite{LuKY23} (Figure 10) shows that the GC attack does not change the semantics of the clean samples much, even without explicit constraints for end-to-end training. However, we show it is not the case for a fixed feature extractor in \Cref{fig:example_image}. Instead, GC input space attack returns noisy samples that lose all information of the input sample. 

Based on the above two observations, we further add two constraints to the two attacks to make them less vulnerable to possible defenses: (1) clipping the poisoned samples to the range of clean samples; (2) adding a term on measuring the (squared) $L_2$ distance between input (clean sample as initialization) and output (poisoned points) specified in \Cref{sec:input_space}. In \Cref{tab:finetuning} (upper part, row 3,5), we find the attacks are much less effective, especially the GC attack, and may not be considered a strong threat. Compared with poisoning end-to-end training in \textcite{LuKY23}, our poisoned samples on fixed feature extractors are much weaker with constraints, e.g., \textcite{LuKY23} report the test accuracy drop decreases $\approx 10\%$ for $\epsilon_d=0.03$, which is significantly less than ours (29.54\% (no constraints) - 2.50\% (with constraints) = 27.04\%). This confirms that with a fixed feature extractor, it is much more difficult to construct feasible poisoning samples than end-to-end training for input space attacks, which motivates us to design and examine stronger attacks.

\begin{table*}[ht]
    \centering
    \caption{Transfer learning results: comparison between input space attacks (TGDA and GC input space attacks with/without constraints on poisoned samples) and feature targeted attacks (Decoder inversion and Feature Matching with different $\beta$) \wrt different $\epsilon_d$. \teal{The white background indicates methods that generate target parameters or features; blue indicates input space attacks and pink indicates feature targeted attacks.}}
    \label{tab:transfer}  
    \setlength\tabcolsep{6pt}
   \resizebox{\textwidth}{!}{\begin{tabular}{lrrrrrr}
\toprule
\bf Attack method & Clean & $\epsilon_d=0.03$ \teal{(1500)} & $\epsilon_d=0.1$ \teal{(5000)} & 
$\epsilon_d=0.5$ \teal{(25000)} & $\epsilon_d=1$ \teal{(50000)}\\

\midrule
GradPC (input space) & 93.03\% & 49.39\% / -43.64\% & 49.39\% / -43.64\% & 49.39\% / -43.64\% & 49.39\% / -43.64\% \\

\rowcolor{LightCyan1}
TGDA-input (no constraints) & 93.03\% & 87.84\% / -\phantom{1}5.19\% & 86.97\% / -\phantom{1}6.06\% & 78.74\% / -14.29\% & 72.93\% / -20.10\% \\

\rowcolor{LightCyan1}
TGDA-input (with constraints)  & 93.03\% & 89.71\% / -\phantom{1}3.32\% & 89.44\% / -\phantom{1}3.59\% & 85.25\% / -\phantom{1}7.78\% & 79.97\% / -13.06\% \\

\rowcolor{LightCyan2}
GC-input (no constraints) & 93.03\% & 53.68\% / -39.35\% & 52.84\% / -40.19\% & 50.04\% / -42.99\%  & 49.48\% / -43.55\% \\

\rowcolor{LightCyan2}
GC-input (with constraints) &  93.03\% & 88.63\% / -\phantom{1}4.40\% & 87.44\% / -\phantom{1}5.59\% & 79.52\% / -\phantom{1}13.51\% & 76.05\% / -16.98\% \\

\midrule

GradPC (feature space) & 93.03\% & 49.53\% / -43.50\% & 49.53\% / -43.50\%  & 49.53\% / -43.50\%  & 49.53\% / -43.50\%  \\

GC-feature space & 93.03\% & 50.01\% / -43.02\% & 50.01\% / -43.02\% & 50.01\% / -43.02\%  & 50.01\% / -43.02\%  \\

\rowcolor{LavenderBlush1}
Decoder inversion & 93.03\% & 87.92\% / -\phantom{1}5.11\%  & 86.96\% / -\phantom{1}6.07\%  & 79.27\% / -13.76\%  & 74.06\% / -18.97\% \\

\rowcolor{LavenderBlush2}
Feature matching  ($\beta=0.25$) & 93.03\% & 71.50\% / -21.53\% & 65.04\% / -27.99\% & 57.93\% / -35.10\% & 50.69\% / -42.34\% \\

\rowcolor{LavenderBlush2}
Feature matching  ($\beta=0.1$) & 93.03\% & 59.58\% / -33.45\% & 57.74\% / -35.29\% & 50.74\% / -42.29\% & 49.63\% / -43.40\% \\

\bottomrule
\end{tabular}}
\vspace{-10pt}
\end{table*}

\begin{figure*}
    \centering   
    \includegraphics[height=5cm]{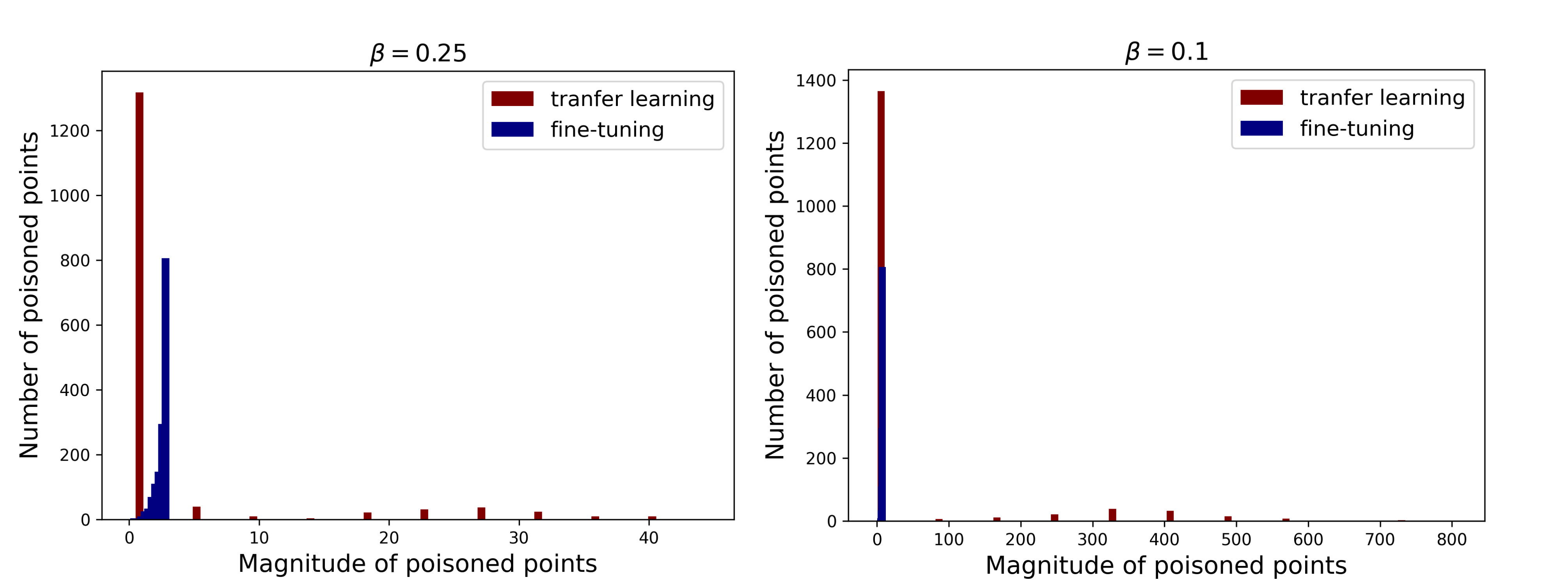}
    \vspace{-10pt}
    \caption{\teal{The histogram of the ($L_{\infty}$) magnitude of 1500 ($\epsilon_d=0.03$) poisoned points generated by feature matching feature matching on transfer learning (red) and fine-tuning (blue) with $\epsilon_d=0.03$ for $\beta=0.25$ (left), $\beta=0.1$ (right). The plot shows that while for fine-tuning the poisoned points are roughly in the same range, for transfer learning, $\approx 10\%$ of the poisoned samples are anomalies. }
    }
    \label{fig:anomalies}
\end{figure*}

\begin{figure*}[t]
    \centering
    \includegraphics[height=6.5cm]{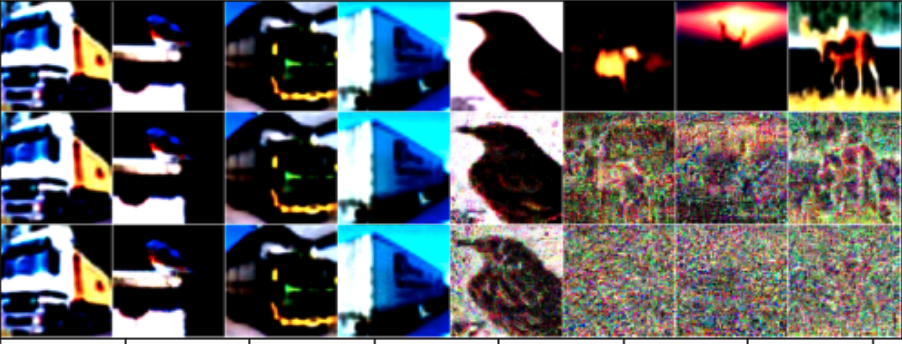}
    \vspace{-5pt}
    \caption{We plot poisoned images generated by feature matching on transfer learning for $\epsilon_d=0.03$. We show clean images and poisoned images for $\beta=0.25$ and $\beta=0.1$ on rows 1,2,3 respectively. Columns 1-4 are images within the normal range, and columns 5-8 are abnormal poisoned images with a bigger magnitude. }
    \label{fig:fm_transfer}
\end{figure*}

\subsubsection{Feature targeted attacks}

Next, we discuss two feature targeted (FT) attacks introduced in \Cref{sec:feature}, namely the decoder inversion and the feature matching attack. Both of the FT attacks involve three stages: (1) GradPC feature space attack for acquiring the target parameters $\hat\omegav$, where we achieve roughly the same performance as that in input space; (2) GC feature space attack for obtaining the target feature $\zeta$; (3) taking $\zeta$ as the input, we either invert the feature back directly using a decoder or perform feature matching. 

\blue{Note that we do not treat GC feature space attack as a viable attack (as injecting poisoned features may not be realistic), but as an essential building block and intermediate step for decoder inversion and feature matching. Additionally, our following results for GC feature space attack provide important information on how good a poisoned feature $\zeta$ can be (in terms of attack performance) and the difficulty of inverting them back to the input space.}

In \Cref{tab:finetuning} (lower part, rows 1-2), we report GradPC (feature space) for stage 1 and GC-feature space for stage 2. We observe that the GC feature space attack is very successful in reproducing the target parameters $\omegav$, even when $\epsilon_d=0.03$. Additionally, comparing with the GC input space attack, we observe that GC feature space induces a much smaller perturbation in terms of magnitude, e.g., the poisoned features are $4\times$ larger than the normal ones for $\epsilon_d=0.03$ and lie within the same range for $\epsilon_d \geq 0.2$.

Surprisingly, although GC feature space attack obtains the target feature $\zeta$ with relatively small perturbations, it is difficult to invert it back to the input space. For decoder inversion (row 3), although we observe that the images are in the normal range and visually benign (see \Cref{fig:u-net}), it is much less effective than the target poisoned features. In practice, we find the poisoned samples are largely dominated by the skip connections of the encoder architecture, and thus the constraint on the input space could be too strong and cannot be easily controlled.

In comparison, the feature matching algorithm (rows 4-6) is more effective than decoder inversion, i.e., for $\beta=0.25$, we observe a further accuracy drop $\approx 2\%$ for every choice of $\epsilon_d$. Moreover, by controlling the hyperparameter $\beta$, we can easily tune the constraints on the input similarity, where a smaller $\beta$ indicates less power of  constraints. In \Cref{tab:finetuning}, we observe that a smaller $\beta$ generally returns a much stronger attack. However, tuning $\beta$ also involves a trade-off: in \Cref{fig:feature_matching}, we visualize the poisoned images generated by feature matching attacks with different $\beta$ and observe that a smaller $\beta$ generally introduce more noise, and poisoned images generated by $\beta=0.1, 0.05$ can be easily discriminated by human eyes. Additionally, smaller $\beta$ induce poisoned images with a larger magnitude, i.e., for $\beta=0.25,0.1,0.05$, we observe the magnitude of $1\times, 3\times, 5.5\times$ the normal range, respectively.

\blue{Moreover, we compare the best-performing input space attack (GC-input) and feature targeted attack (feature matching). Recall that both of these methods aim to solve the same constrained problem in \Cref{eq:gc_input}, where GC-input directly the constrained gradient canceling problem while the latter approach mitigates the difficulty of the constrained problem with the staged strategy in \Cref{eq:gc_input-relax}. We observe that the feature matching attack indeed helps with the optimization procedure and improves the attack effectiveness by a big margin (range from 3.80\% to 6.16\% for different $\epsilon_d$).} 

In summary, we find both of the feature targeted attacks consistently outperform input space attacks across all choices of $\epsilon_d$. However, for indiscriminate attacks in general, inverting poisoned features back to the input space attack is rather difficult, and cannot be achieved exactly with our existing attacks.

\subsection{Transfer Learning}

Next, we present our experimental results for the task of transfer learning. Recall that for transfer learning, we take the feature extractor (ResNet-50) pre-trained on ImageNet without labels and perform a linear evaluation on the CIFAR-10 dataset with labels. We report our main results in \Cref{tab:transfer} and expand our discussions as follows.

\subsubsection{Input space attacks}

Firstly we perform the same input space attacks on the transfer learning task. In \Cref{tab:transfer} (upper part), we observe: (1) In general, input space attacks are more effective (an increased accuracy drop $\approx$  2\%-10\%) on transfer learning than on fine-tuning tasks. A large factor could be that transfer learning involves a distribution shift, such that the linear evaluation process is more vulnerable to poisoning attacks; (2) without any constraints, GC input space still outperforms TGDA input space. Meanwhile, GC generates poisoned samples that have a larger magnitude than that of fine-tuning, i.e., about $290\times$ larger for $\epsilon_d=0.03$ (fine-tuning: $100\times$ larger) than clean samples, and reduces to $14\times$ larger for $\epsilon_d=1$ (fine-tuning: $5\times$ larger). As for poisoned samples, TGDA and GC input space attacks still generate noisy images that have no semantic meanings (analogous to that in \Cref{fig:example_image}); (3) again we add the same constraints on magnitude and visual similarity with the input images and observe that the attack performance significantly drops.

\begin{table}[t]
    \centering
    \caption{Comparison between feature matching and feature matching after removing anomaly points (feature matching-r).}
    \label{tab:fm-r}  
    \setlength\tabcolsep{6pt}
   \resizebox{0.48\textwidth}{!}{\begin{tabular}{lrrr}
\toprule
\bf Attack method & Clean & $\epsilon_d=0.03$ \\

\midrule

Feature matching\phantom{-r}  ($\beta=0.25$) & 93.03\% & 71.50\% / -21.53\% \\

Feature matching-r  ($\beta=0.25$) & 93.03\% & 85.82\% / -\phantom{1}7.21\% \\

\midrule

Feature matching\phantom{-r}  ($\beta=0.10$) & 93.03\% & 59.58\% / -33.45\%  \\

Feature matching-r  ($\beta=0.10$) & 93.03\% & 86.38\% / -\phantom{1}6.65\% \\

\bottomrule
\end{tabular}}
\vspace{-10pt}
\end{table}

\subsubsection{Feature targeted attacks}
We perform the same feature targeted attacks and report the results in \Cref{tab:transfer} (lower part). For the feature matching attacks, we omit the results for $\beta=0.05$ as it does not show a substantial improvement over $\beta=0.1$. We observe that in general, similar to input space attacks, the attacks are more effective on transfer learning tasks than on fine-tuning. Additionally, decoder inversion and feature matching still outperform all input space attacks considering constraints across all choices of $\epsilon_d$. 

Notably, upon further exploration, we observe that feature matching behaves differently in transfer learning. Recall that for fine-tuning when we choose $\beta=0.25$, the magnitude of poisoned samples is within the normal range, and for $\beta=0.1$, the magnitude is generally $3\times$. However, for transfer learning, the algorithm returns several abnormal images with a larger magnitude and more noise. In \Cref{fig:anomalies}, we randomly select 100 poisoned samples (out of 1500 for $\epsilon_d=0.03$) and plot their $L_{\infty}$ magnitude. For the entire poisoned set, we observe that for $\beta=0.25$ and $\beta=0.1$, there are 181 (12.1\%) and 274 (18.2\%) anomalies, respectively. Additionally, when $\beta$ is smaller, while most of the poisoned points are in the normal range, the anomalies have a much bigger magnitude, due to the power of the constraint being weakened. In \Cref{fig:fm_transfer}, we compare poisoned images within the normal range (columns 1-4) and abnormal ones (columns 5-9). We observe that abnormal poisoned images also contain more noise, especially when $\beta$ is smaller.
To examine the real effect of feature matching attacks, we perform a simple data sanitization process by removing all of the abnormal points before evaluation. We report the results in \Cref{tab:fm-r} and observe the attack becomes much less effective after the defense. Nevertheless, feature matching-r ($\beta=0.25$) still outperforms other baseline attacks.  

Overall, compared to fine-tuning, we find transfer learning easier to poison for every attack, with or without constraints in the input space. However, inverting poisoned features back to input space, either explicitly through feature targeted attacks or implicitly through input space attacks, is still difficult in the condition of considering poisoned points in a feasible set. Specifically, we find that no existing attack can exactly reproduce the target parameters (features) with a limited number of visually benign poisoned samples.

\begin{table}[t]
    \centering
    \caption{Comparison between UE (EMN) in the context of poisoning a fixed feature extractor with linear head and training end-to-end (with the same initialization). We report the results on both fine-tuning and transfer learning.}
    \label{tab:ue}  
    \setlength\tabcolsep{6pt}
   \resizebox{0.48\textwidth}{!}{\begin{tabular}{llrrr}
\toprule
\bf Task & \bf Attack & Clean & $\epsilon_d=\infty$ \\

\midrule

\multirow{3}{*}[-.6ex]{\bf Fine-tuning} &
Original & 94.77\% & 19.93\% / -74.84\% \\

& End-to-end   & 95.22\% & 68.79\% / -26.43\% \\

& Input space  & 89.71\% & 82.16\% / -\phantom{1}7.55\% \\

\midrule

\multirow{3}{*}[-.6ex]{\bf Transfer Learning} &
Original & 94.42\% & 18.89\% / -75.53\% \\

& End-to-end  & 95.09\% & 66.05\% / -29.04\% \\

& Input space  & 93.03\% & 85.15\% / -\phantom{1}7.88\% \\

\bottomrule
\end{tabular}}
\end{table}

\begin{figure*}[t]
    \centering
    \includegraphics[height=7cm]{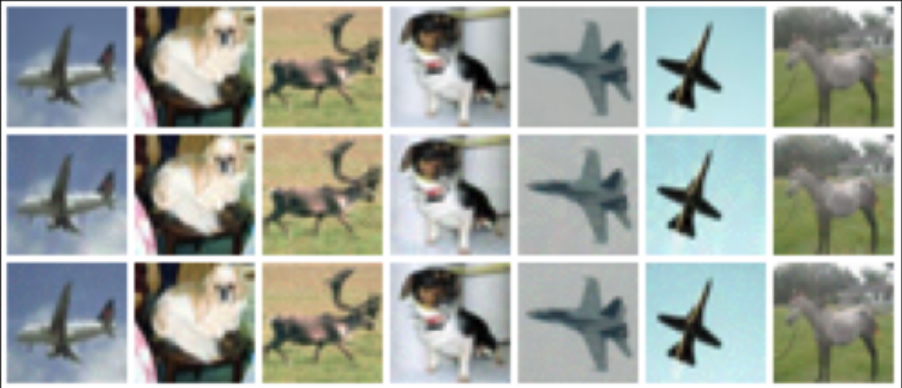}
    \vspace{-5pt}
    \caption{We plot poisoned images generated by EMN input space attack. We show clean images and poisoned images for fine-tuning (which incurs an accuracy drop of 7.55\%) and transfer learning (incurring an accuracy drop of 7.88\%) on rows 1,2,3, respectively.}
    \label{fig:ue}
    \vspace{-10pt}
\end{figure*}
\subsection{Unlearnable Examples}

Finally, we discuss the unlearnable examples of poisoning fixed feature extractors (the EMN input space attack). We apply the attack of \textcite{HuangMEBW21} and report the results in \Cref{tab:ue}. We compare three variants of EMN: (1) original: we consider the original setting in \textcite{HuangMEBW21}, where during the attack and evaluation, the entire model is updated with random initialization on the unlearnable dataset\footnote{The results are directly obtained from Table 1 in \textcite{HuangMEBW21}, CIFAR-10 ResNet-18/50.}; (2) end-to-end: the model is initialized with the pre-trained feature extractor, but all parameters can be updated; (3) input space: we consider our setting where the feature extractor is fixed during training and evaluation. We visualize the poisoned images generated by EMN in \Cref{fig:ue}.

From \Cref{tab:ue} we obtain several interesting observations: (1) EMN is highly sensitive to weights initialization: by comparing original and end-to-end, we show that random initialization is much more vulnerable to EMN than a good initialization (i.e., pre-trained weights by contrastive learning); (2) EMN input space is not very effective: with a significantly larger poisoning budget $\epsilon_d=\infty$, EMN only performs similar to feature matching when $\epsilon_d=0.03$. Compared with original and end-to-end, EMN input space is also less than a magnitude effective for both fine-tuning and transfer learning.

In summary, our observations reveal that training a neural network architecture with a good feature extractor as initialization, especially when it is frozen, is much more robust to the EMN attack.

%% file: rebuttal.tex
\begin{table*}[ht]
    \centering
    \caption{\teal{Transfer learning results on vision transformers: comparison between input space attacks (GC input space attacks with/without constraints on poisoned samples) and feature matching after removing anomaly points (feature matching-r) \wrt different $\epsilon_d$. }}
    \label{tab:vit}  
    \setlength\tabcolsep{6pt}
   \resizebox{\textwidth}{!}{\begin{tabular}{lrrrrrr}
\toprule
\bf Attack method & Clean & $\epsilon_d=0.03$ \teal{(1500)} & $\epsilon_d=0.1$ \teal{(5000)} & 
$\epsilon_d=0.5$ \teal{(25000)} & $\epsilon_d=1$ \teal{(50000)}\\

\midrule
GradPC (input space) & 98.90\% & -46.53\% & -46.53\% & -46.53\% & -46.53\% \\

GC-input (no constraints) & 98.90\% & -40.12\% & -42.30\% & -44.66\% &   -46.49\% \\

GC-input (with constraints) &  98.90\% & -\phantom{1}4.28\% & -\phantom{1}5.99\% & -\phantom{1}13.12\% & -17.66\% \\

\midrule

GradPC (feature space) & 98.90\% & -46.53\% & -46.53\% & -46.53\% & -46.53\% \\

GC-feature space & 98.90\% & -46.51\% & -46.51\% & -46.51\% & -46.51\%  \\

Feature matching-r  ($\beta=0.25$) &  98.90\% & -\phantom{1}6.33\% & -\phantom{1}7.89\% & -\phantom{1}15.04\% & -19.27\%\\

\bottomrule
\end{tabular}}
\end{table*}

\section{Perturbations introduced by feature matching}

\begin{figure}[ht]
\centering\includegraphics[height=3.5cm]{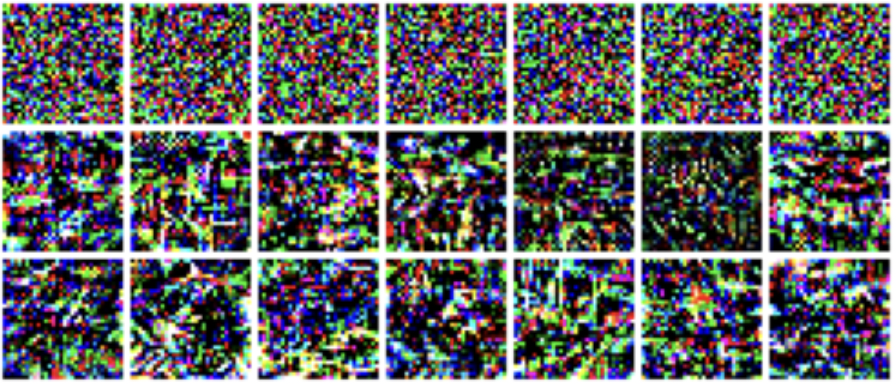}
    \caption{Here we visualize perturbations introduced by the feature matching algorithm with $\beta=0.25, 0.1, 0.05$ respectively from the first to the third row.}
    \label{fig:feature_matching_diff}
\end{figure}

\teal{Recall in \Cref{fig:feature_matching}, we visualized the poisoned images generated by feature matching attacks with different $\beta$. Here in \Cref{fig:feature_matching_diff} we also visualize the difference between the clean images and the poisoned images to show the perturbation. }

\section{Additional results on transfer learning}

\subsection{Vision transformer as encoder}\teal{Aside from the ResNet architecture, another popular backbone architecture serving as a feature extractor is the recently advanced vision transformers \parencite{DosovitskiyBKWZUDMHGS20}. Here we extend our experiments of poisoning transfer learning on pre-trained vision transformers (specifically, ViT-base) with MoCo V3 \parencite{ChenXH21} on ImageNet to CIFAR-10. We report our results in \Cref{tab:vit} and observe that our conclusion still holds for the ViT architecture. }

\subsection{Image encoder trained with CLIP} 
\teal{In this paper, we introduce contrastive learning methods with one modality, i.e., images. Another popular approach to learning feature extractors considers both images and texts, also known as contrastive language-image pre-training (CLIP) \parencite{RadfordKHRGASAMC21}. Instead of choosing different augmentations of the same image as positive pairs and different images as negative pairs, CLIP considers labeled image-text pairs as positive ones, and unmatched pairs as negative ones. Specifically, CLIP trains an image encoder and a text encoder simultaneously, which can be used separately for downstream tasks. Here we examine data poisoning attacks against the image encoder trained by CLIP with a ResNet-50 architecture. We directly adopt the pre-trained weights acquired by OpenAI \footnote{\url{https://github.com/openai/CLIP}}. We report our result in \Cref{tab:clip} and observe that feature matching still outperforms GC input space attack (with constraints), even after removing anomaly points.
}

\begin{table}[hbt!]
    \centering
    \caption{\teal{Transfer learning results on CLIP image encoder: comparison between input space attacks (GC input space attacks with/without constraints on poisoned samples) and feature matching after removing anomaly points (feature matching-r).}}
    \label{tab:clip}  
    \setlength\tabcolsep{6pt}
   \resizebox{0.5\textwidth}{!}{\begin{tabular}{lrrr}
\toprule
\bf Attack method & Clean & $\epsilon_d=0.03$ \teal{(1500)} \\

\midrule
GradPC (input space) & 92.30\% & -44.55\% \\

GC-input (no constraints) & 92.30\% & -41.20\% \\

GC-input (with constraints) &  92.30\% & -\phantom{1}4.39\% \\

\midrule

GradPC (feature space) & 92.30\% & -44.51\% \\

GC-feature space & 92.30\% & -42.99\% \\

Feature matching-r  ($\beta=0.25$) &  92.30\% & -\phantom{1}7.05\% \\

\bottomrule
\end{tabular}}
\end{table}

\subsection{Transferring to CIFAR-100}

\teal{In \Cref{tab:transfer}, we show the results of transferring a pre-trained feature extractor (trained on ImageNet) to the CIFAR-10 dataset. As CIFAR-10 only involves 10 classes, the classification problem is rather easy. Here we extend the downstream task to a more difficult setting with 100 classes, i.e. the CIFAR-100 dataset \parencite{Krishevsky09}. Specifically, CIFAR-100 consists of 100 classes containing 600 images each, with 500 training images and 100 testing images respectively. Again for ImageNet pre-training we apply the ResNet-50 architecture using MoCo V3 \parencite{ChenXH21} and examine the performance of data poisoning attacks against transferring to CIFAR-100. We report the results in \Cref{tab:cifar-100} and observe that: (1) transferring to CIFAR-100 is more challenging than CIFAR-10 (79.90\% clean accuracy compared to CIFAR-10, 93.03\%); (2) the attacks are generally weaker under a fine-grained classification setup; (3) feature matching-r still outperform GC input space (with constraints). }

\begin{table}[hbt!]
    \centering
    \caption{\teal{Transfer learning results on CIFAR-100: comparison between input space attacks (GC input space attacks with/without constraints on poisoned samples) and feature matching after removing anomaly points (feature matching-r).}}
    \label{tab:cifar-100}  
    \setlength\tabcolsep{6pt}
   \resizebox{0.5\textwidth}{!}{\begin{tabular}{lrrr}
\toprule
\bf Attack method & Clean & $\epsilon_d=0.03$ \teal{(1500)} \\

\midrule
GradPC (input space) & 79.90\% & -38.51\% \\

GC-input (no constraints) & 79.90\% & -36.78\% \\

GC-input (with constraints) &  79.90\% & -\phantom{1}3.29\% \\

\midrule

GradPC (feature space) & 79.90\% & -38.50\% \\

GC-feature space & 79.90\% & -37.99\% \\

Feature matching-r  ($\beta=0.25$) &  79.90\% & -\phantom{1}6.69\% \\

\bottomrule
\end{tabular}}
\end{table}

\begin{figure}
    \centering\includegraphics[height=2.5cm]{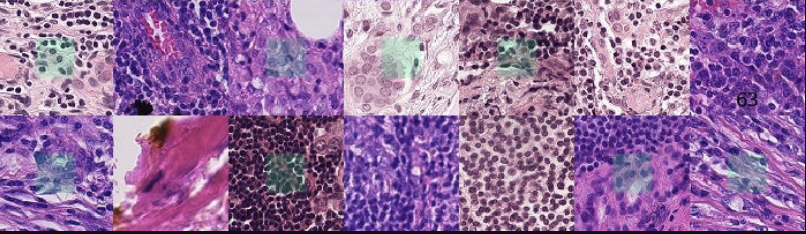}
    \caption{Example images from the PatchCamelyon dataset.}
    \label{fig:patchcam}
\end{figure}

\subsubsection{Transferring to PatchCamelyon}
\teal{Next we consider a dataset with a more drastic domain shift from the pre-training dataset, i.e., ImageNet. Here we choose the PatchCamelyon dataset \parencite{VeelingLWCW18,Bejnordietal17}, which consists of 327680 (220025 training samples) color images of size $96\times96$, extracted from histopathologic scans of lymph node selection. Each image is annotated with a binary label indicating the presence of metastatic tissue. In \Cref{fig:patchcam}, we visualize some samples in the dataset. Again, we choose the feature extractor pre-trained by MoCo V3 \parencite{ChenXH21} (ResNet-50) on ImageNet and examine the performance of data poisoning attacks against transferring to PatchCamelyon. Note that we follow \parencite{TruongML21} and randomly choose 5000 samples from the original dataset for the downstream training. We report our results in \Cref{tab:patchcam} and observe that transferring to PatchCamelyon is generally more vulnerable to data poisoning attacks (higher accuracy drop) than CIFAR-10/100, which may indicate that a drastic domain shift is more fragile against such attacks.
}

\begin{table}[hbt!]
    \centering
    \caption{\teal{Transfer learning results on PatchCamelyon: comparison between input space attacks (GC input space attacks with/without constraints on poisoned samples) and feature matching after removing anomaly points (feature matching-r).}}
    \label{tab:patchcam}  
    \setlength\tabcolsep{6pt}
   \resizebox{0.5\textwidth}{!}{\begin{tabular}{lrrr}
\toprule
\bf Attack method & Clean & $\epsilon_d=0.03$ \teal{(150)} \\

\midrule
GradPC (input space) & 79.12\% & -50.13\% \\

GC-input (no constraints) & 79.12\% & -49.05\% \\

GC-input (with constraints) &  79.12\% & -11.03\% \\

\midrule

GradPC (feature space) & 79.12\% & -50.12\% \\

GC-feature space & 79.12\% & -50.00\% \\

Feature matching-r  ($\beta=0.25$) &  79.12\% & -15.33\% \\

\bottomrule
\end{tabular}}
\end{table}